\documentclass[10pt,journal]{IEEEtran}
\usepackage{times}
\usepackage{epsfig}
\usepackage{graphicx}
\usepackage{amsmath}
\usepackage{amssymb}
\usepackage[ruled]{algorithm2e}
\usepackage{algpseudocode}
\usepackage{caption}
\usepackage{subfigure}
\usepackage{booktabs}
\usepackage{tabularx}
\usepackage{xcolor}
\usepackage{multirow}
\usepackage{cite}
\usepackage{hyperref}

\newcolumntype{Y}{>{\centering\arraybackslash}X}

\begin{document}
%

\title{MetaFi: Device-Free Pose Estimation via Commodity WiFi for Metaverse Avatar Simulation}

\author{%
  \IEEEauthorblockN{%
    Jianfei Yang,
    Yunjiao Zhou,
    He Huang,
    Han Zou and
    Lihua Xie%
  }%
  
   School of Electrical and Electronics Engineering, Nanyang Technological University, Singapore   \\

  Email: \{yang0478,zh0027ao,he008,zouh0005,elhxie\}@ntu.edu.sg
}

\markboth{}%
{Shell \MakeLowercase{\textit{et al.}}: Bare Demo of IEEEtran.cls for IEEE Journals}

\maketitle

\begin{abstract}
   Avatar refers to a representative of a physical user in the virtual world that can engage in different activities and interact with other objects in metaverse. Simulating the avatar requires accurate human pose estimation. Though camera-based solutions yield remarkable performance, they encounter the privacy issue and degraded performance caused by varying illumination, especially in smart home. In this paper, we propose a WiFi-based IoT-enabled human pose estimation scheme for metaverse avatar simulation, namely MetaFi. Specifically, a deep neural network is designed with customized convolutional layers and residual blocks to map the channel state information to human pose landmarks. It is enforced to learn the annotations from the accurate computer vision model, thus achieving cross-modal supervision. WiFi is ubiquitous and robust to illumination, making it a feasible solution for avatar applications at smart home. The experiments are conducted in the real world, and the results show that the MetaFi achieves very high performance with a PCK@50 of 95.23\%.
\end{abstract}


\section{Introduction}
Metaverse refers to a synthetic environment that is mapped from the physical world. Enabled by the rapid development of Computer Graphics (CG), Human-Computer Interaction (HCI), Internet of Things (IoT) and Artificial Intelligence (AI), the metaverse era has come and many related applications are emerging. In metaverse, a virtual environment parallel to the real world, digital avatar is a vital technology that enables users to interact with the virtual world synchronously. Each user in metaverse will have a customized avatar and experience an alternate life in a virtuality~\cite{lee2021all}. 

To create a synchronous digital avatar, human pose estimation captures 2D or 3D human landmarks that are used as the human skeleton by CG developers. Current approaches for human pose estimation mainly rely on cameras equipped with computer vision algorithms. Supported by novel deep neural networks, visual pose estimation has achieved very high accuracy and facilitated avatar digitization~\cite{li20193d}. Nevertheless, these methods still encounter difficulties in practical applications. Firstly, occlusion can degrade the performance of visual algorithms. Restricted by the range of camera, occlusion and self-occlusion could be inevitable in small spaces. Secondly, visual pose estimation suffers from bad illumination condition that is commonly seen at night. Thirdly, in smart home metaverse, camera-based solutions can arouse severe privacy issue. Regarding these challenges, a cost-effective privacy-preserving noise-robust solution for pose estimation is highly demanded in digital avatar simulation at smart home.

\begin{figure}[t]
	\centering
	\includegraphics[width=0.88\linewidth]{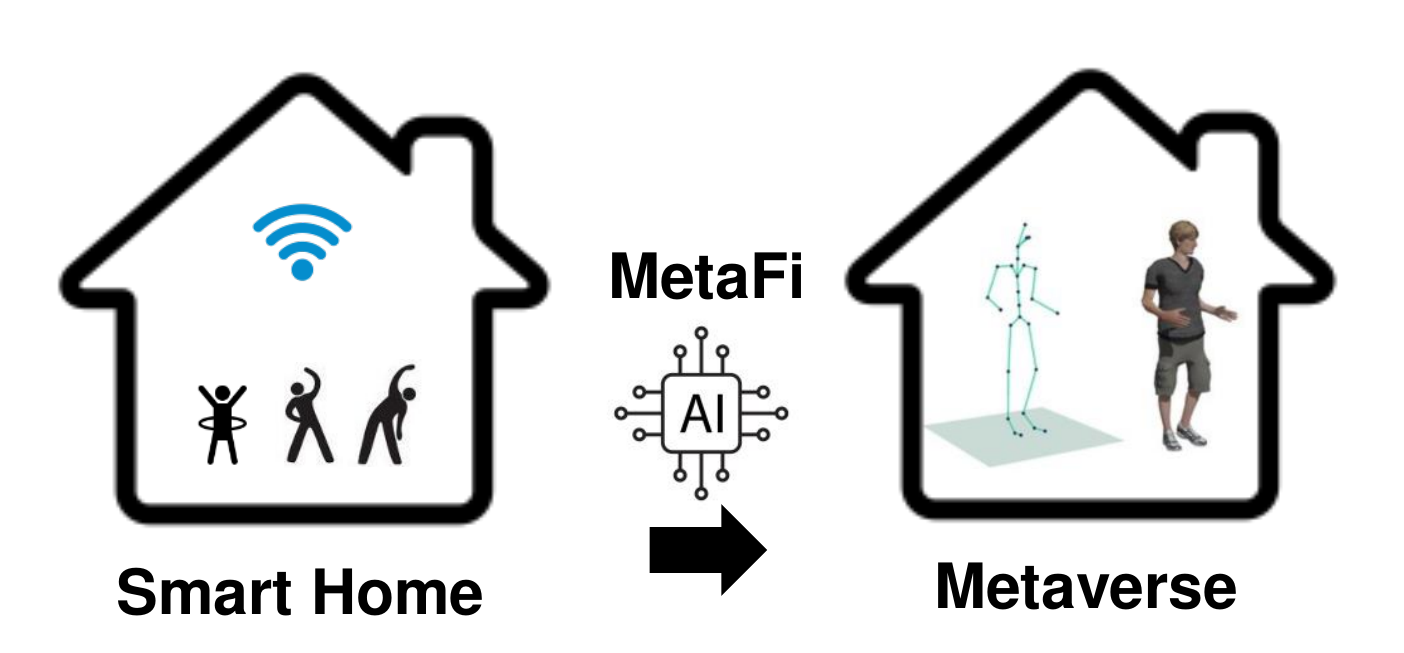}
	\caption{Illustration of the MetaFi that connects smart home and metaverse using WiFi.}
	\label{fig:principle}
\end{figure}

Recently, WiFi sensing has attracted great attention since it is cost-effective and simultaneously preserves user privacy. In smart home and buildings, WiFi has served as a ubiquitous infrastructure for Internet access and indoor localization~\cite{zou2017non,zou2018unsupervised}. Previous researches have shown that human pose can affect the propagation of WiFi signals and these signal variations are reflected by Channel State Information (CSI). Enabled by advanced deep learning methods, many studies have been conducted to enable various applications, such as human activity recognition~\cite{zou2018deepsense,yang2018carefi,zou2019wificv}, gesture recognition~\cite{yang2019learning,zou2018gesture}, human identification~\cite{zou2018identification,wang2022caution}, and people counting~\cite{zou2018device,FreeCount}. Many advanced learning methods further promote the automation and performance of these tasks, such as transfer learning~\cite{zou2018robust,yang2020mobileda}, unsupervised learning~\cite{yang2022autofi}, and model-robust learning~\cite{yang2022robustsense}. Then we imagine that if commodity WiFi is able to recognize human pose, it will be an ideal solution for metaverse avatar simulation in smart home.

To this end, we study whether commercial off-the-shelf (COTS) WiFi devices are capable of estimating human poses. Though some previous works have similar goals~\cite{jiang2020towards}, they rely on laptops with specific network cards~\cite{halperin2011tool} and expensive annotation systems. In this paper, we only leverage COTS WiFi devices for pose estimation. Our method, namely MetaFi, relies on cross-modal supervision between WiFi and camera at the training stage, and enables pure WiFi devices to estimate human pose. Specifically, we develop an IoT-enabled WiFi sensing system to extract CSI data, and obtain the annotations by an accurate visual pose estimation method. Then a deep neural network is customized with mostly convolutional layers to regress the pose landmarks. The model is trained using the supervision from the image modality, i.e., the synchronous landmarks estimated by computer vision methods. The human pose landmarks are used for avatar simulation in metaverse. Realistic experiments show that our method achieves high accuracy. 

We summarize the contributions as follows:
\begin{itemize}
    \item We leverage COTS WiFi for human pose estimation, which is the first solution based on wireless sensing for metaverse avatar simulation.
    \item We propose a customized deep neural network that conducts cross-modal supervision and learns to estimate human poses.
    \item The real-world experiments are conducted and the results demonstrate that MetaFi can generate accurate human pose landmarks.
\end{itemize}


\section{Related Work}
\subsection{Metaverse Avatar} 
Digital human, also known as avatar in metaverse, has always been the core from virtual worlds in computer games to the emerging metaverse applications, e.g., enhanced social media experience, immerse sales and marketing, etc~\cite{ning2021survey}. 
Avatar is the representative of a physical user in the virtual world that is capable of engaging in different activities and interacting with other digital objects including NPCs, virtually constructed environments and other connecting users~\cite{heo2021effect, zhao2022metaverse}. 
Features of physical users including facial expression, body movement and even psychological status are vital for metaverse avatar creation~\cite{zhao2022metaverse}. 
Recently, related works have been proposed to develop an avatar that mimics the physical users. 
In social media applications, users always prefer the avatar that embodies the unique signature to themselves and is more engaged~\cite{vugt2008effects}. 
In~\cite{park2021individual}, the authors studied how the facial habitual expressions and facial appearances in virtual avatars affect social perceptions. 
This study revealed that the participants are more likely to perceive similarities to virtual avatars that have similar facial expressions. 
In education area, the virtual avatar can also improve the effectiveness for novices~\cite{grubert2016towards}. 
In another work~\cite{heo2021effect}, the authors designed a new AR-based sport climbing system that helps guide the learners.
The postures and motions of an instructor are mapped to a virtual avatar in advance, which is further projected onto the artificial climbing structure. 
As a result, the learner can train repeatedly without the presence of instructors, which will be more free and affordable. 

\subsection{Pose Estimation} 
For the development of metaverse avatars, motion estimation plays an important role since it can not only reflect the body status, but also serve as clues for emotion analysis~\cite{zhao2022metaverse}. 
Human pose estimation usually relies on vision sensing, i.e. recognizing poses from images, and can be divided into two groups: two-step framework and part-based framework~\cite{li2018crowdpose}. 
The popular two-step framework includes human bounding box detection and pose estimation within the detected boxes.
One challenge for such framework is the inaccurate bounding box detection that degrades the recognition performance~\cite{li2018crowdpose}. 
AlphaPose was proposed to deal with the inaccurate problems, which consists of a symmetric spatial transformer network (SSTN), parameter pose non-maximum-suppression (NMS) and pose-guided proposal generator (PGPG)~\cite{fang2017rmpe}. 
The SSTN is used to extract high accuracy single person pose estimators even when inaccurate boxes are given. 
The NMS applies the pose similarity comparison method to reduce redundant poses produced in detection. 
For small-scale samples, the PGPG learns the output distribution of human boxes and thus augments the training data. 
Besides the Alphapose, another pose estimation algorithm named OpenPose was proposed in~\cite{cao2019openpose}. 
Different from AlphaPose, OpenPose applies the Part Affinity Field (PAF) to learn the associated body parts in the images. 
The locations of multiple human bodies are predicted on a confidence map, and then the degree of association between parts is encoded in a PAF, which is processed by greedy inference together to generate keypoints for all people in the image. 
OpenPose can achieve comparable performance in contrast to AlphaPose. 
One big advantage of OpenPose is the computational cost since there is no bounding box detection and single person pose estimation. 

\subsection{WiFi-based Human Sensing} 

Despite the popular vision sensing, WiFi-based sensing has also attracted increasing attention in recent years~\cite{zou2020adversarial}. 
WiFi sensing is usually categorized into two parts: Received Signal Strength (RSS) and Channel State Information. 
WiFi RSS has been widely used as an alternate solution for GPS in indoor localization due to the ubiquitous access in indoor environments and popularity of smart devices equipped with network cards~\cite{zou2020adversarial}. 
In contrast to RSS, CSI contains more contextual information since the objects in an environment will affect the propagation of wireless signals~\cite{zhang2021csi}. 
As a result, the variations in amplitude and phase of different channels are regarded as a new type of pattern, which can be used for recognition and inference. 

Recent years have witnessed the emerging applications of human activity recognition (HAR) based on WiFi CSI. 
The authors in~\cite{chen2018wifi} proposed an RNN-based approach, namely ABLSTM, to learn representative features from raw CSI sequence measurements by the utilization of a two-directional Long Short-Term Memory (LSTM). 
Also, an attention mechanism was incorporated to assign different weights according to those learned features. 
As a result, new weighted features are used for final activity recognition model. 
However, many deep models cannot be deployed on edge devices due to the limited computation resources. 
To enable edge computing for CSI sensing, \cite{yang2020mobileda} proposed a new framework named MobileDA for human activity recognition with edge computing adaptability. 
In \cite{yang2020mobileda}, the large training dataset is processed by the teacher network on the cloud server, whereas the testing dataset is used to train the student network on the edge device via cross-domain knowledge distillation. 
In the teacher-student framework, the teacher network is trained in the first place for knowledge learning in the source domain, which is used to produce pseudolabels for student network training in the target domain. 
Then, the student network is trained under the supervision of the teacher network by using the pseudolabels. 
Since environmental changes bring considerable impacts on WiFi CSI data, traditional CSI-based models lack the generalization ability for a new environment. 
In \cite{wang2021multimodal}, MCBAR, a multimodal CSI-based activity recognition approach, is proposed to deal with CSI data diversity in different environments. 
A translation generator is applied to decompose the latent features into two parts: content and interference (environmental impact). 
The content features from different domains should be aligned for consistency, which is used to train the general classifier model for all domains. 

The utilization of HAR has proved the effectiveness of WiFi CSI sensing for activity classification. 
However, seldom research has been conducted to study whether WiFi CSI can be further used for higher resolution activity recognition, i.e. human pose estimation. 
Thus, in this work, we will investigate the potentiality of WiFi CSI-based human pose estimation for a future-proofing metaverse sensing solution without violating user privacy. 
\section{Main Work}
\subsection{MetaFi System Design}
The MetaFi system consists of two COTS WiFi routers (i.e., TP-Link N750) that are utilized as transmitter and receiver. The camera and the receiver are placed in parallel with a height of 1 meter. We upgrade the openwrt systems of two WiFi routers using the IoT-enabled real-time system~\cite{yang2018device}. The transmitter continues to send packages to the receiver, and the high-quality sensing data is collected at the receiver side. Such data is then transmitted to a powerful server for AI model inference, and then the estimated human pose landmarks will be reported via API for Metaverse avatar simulation.

To begin with, we introduce the IoT-enabled WiFi system for pose estimation. In wireless communication, CSI reports the channel properties of a communication link, which describes how WiFi signals propagate in a physical environment with the multi-path effect and physical changes, such as reflections, diffraction and scattering. Following the IEEE 802.11 standard, current WiFi devices employ Orthogonal Frequency Division Multiplexing (OFDM) at the physical layer, and thus multiple antennas are enabled. For each pair of antennas, the amplitude attenuation and phase shift of multi-paths on each communication subcarrier are recorded by CSI data. Thus, it is able to reflect human motions that affect the propagation of WiFi signals. The WiFi signals are modeled as Channel Impulse Response (CIR) $h(\tau)$ in frequency domain:
\begin{equation}
h(\tau)=\sum_{l=1}^{L}\alpha_l e^{j\phi_l} \delta(\tau-\tau_l),
\end{equation}
where $\alpha_l$ and $\phi_l$ represent the amplitude and phase of the $l$-th multi-path component respectively, $\tau_l$ is the time delay, $L$ indicates the total number of multi-path and $\delta(\tau)$ denotes the Dirac delta function. Based on the CSI tool (e.g., the Atheros Tool~\cite{xie2015precise}), the OFDM receiver samples the signal spectrum at subcarrier level, which comprises amplitude attenuation and phase shift via complex numbers. The estimation of CSI is denoted by
\begin{equation}
H_i=||H_i||e^{j \angle H_i},
\end{equation}
where $||H_i||$ and $\angle H_i$ are the amplitude and phase of $i$-th subcarrier, respectively. Compared to the prevailing Intel 5300 NIC that can only obtain 30 subcarriers of CSI, our system has 114 subcarriers of CSI for each pair of antennas. In MetaFi, we leverage 3 antennas at the receiver side and 1 antenna at the transmitter side, and only use the amplitude data, as the phase shift is quite noisy. Finally, the CSI data has a size of $3\times 114\times T$ where $T$ is the package number of a specific duration. We synchronize the WiFi CSI and vision sensing to make one video frame correspond to a bunch of CSI frames with $T=32$. 

Let us denote the CSI frame and video frame by $x_c$ and $x_v$, respectively. We use the HRNet~\cite{sun2019deep} in MMPose~\cite{mmpose2020} for each video frame to generate the 17 pose landmarks, used as the annotation. The 17 pose landmarks consist of the coordinates of pixels $(a,b)$ in the image, written as $\mathbf{y}=\{(a_i,b_i)| i\in[1,17]\}$. The objective of our MetaFi system is to estimate the pose landmarks $\hat{\mathbf{y}}$ using only WiFi CSI input $x_c$.

\begin{figure}[t]
	\centering
	\includegraphics[width=1.0\linewidth]{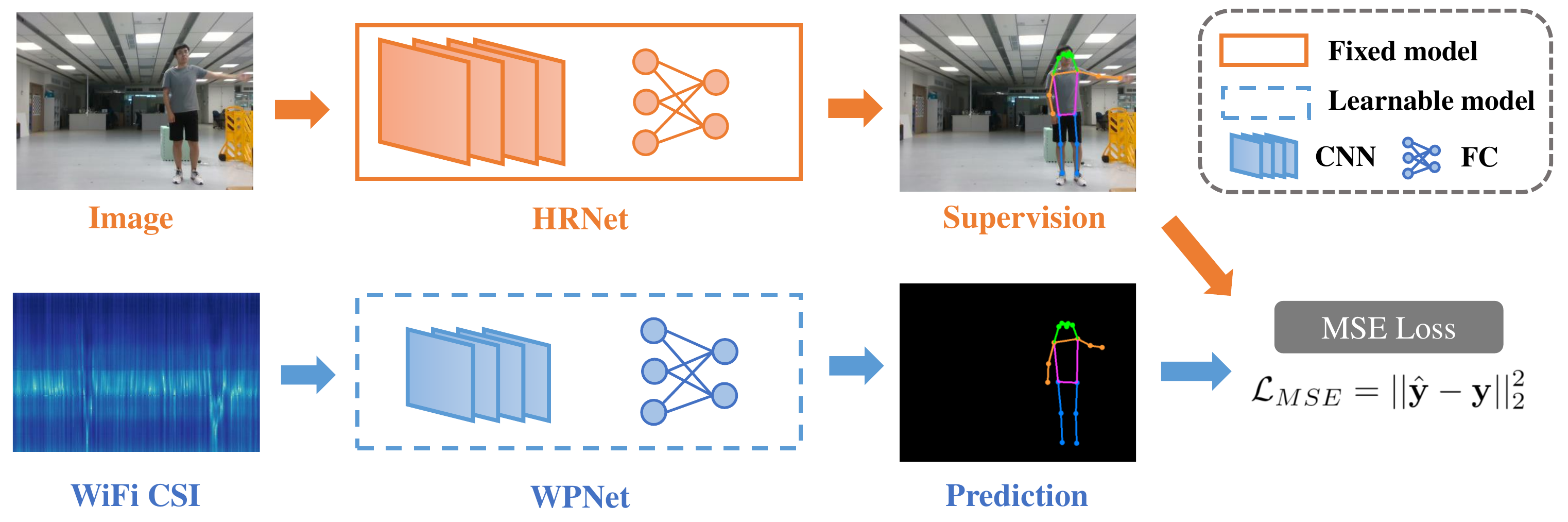}
	\caption{Illustration of the MetaFi framework. The HRNet processes the video frame and extracts the pose annotation that is leveraged for cross-modal supervision of the WPNet.}
	\label{fig:framework}
\end{figure}

\subsection{WPNet: WiFi Pose Network}
For the synchronized video frame and CSI frame, the MetaFi aims to transfer knowledge from the computer vision model to the WiFi model so that the WiFi model can learn to estimate human poses. As shown in Figure~\ref{fig:framework}, our WiFi-based pose estimation model is based on the prevailing deep learning. Firstly, the HRNet~\cite{sun2019deep} generates the pose landmarks $\mathbf{y}$. Then we design the WiFi Pose Network (WPNet) to map the WiFi CSI data to regress the human poses. The WPNet consists of a convolutional neural network as the feature extractor and a bottleneck layer. The learning objective is to minimize the error between the prediction of WPNet and the result of HRNet, which is a cross-modal supervision process.

When the CSI data $x_c\in \mathbb{R}^{3\times114\times32}$ is collected, we firstly preprocess the data to fit the network input. Similar to images, to prevent irregular convolutional kernels, we resize the CSI data to a square matrix. We resize the CSI frame to $114\times(3\times32)=114\times96$, and then upsample the frame to a $1\times136\times136$ tensor by interpolation. The first 136 dimension is the subcarrier axis that contains how signal propagates across the space, i.e., spatial information~\cite{yang2018device}, while the second 136 dimension consists of the antenna and temporal information. The 2D convolutional layer can capture the features across space and time with small parameter size, so it is chosen as the backbone layer in WPNet.

Inspired by the experience of state-of-the-art network design in computer vision, we design the WPNet using convolutional layers~\cite{lecun1998gradient} and residual block~\cite{he2016deep}, as shown in Table~\ref{tab:network}. In Table~\ref{tab:network}, each block consists of a convolutional kernel of $3\times3$ with $c$ channels. The residual block is established every 2 layers in the 2-5th blocks. The output size of each block is shown in the middle column. The last block generates a tensor size of $512\times17\times17$. Then we innovatively use a bottleneck layer with 2 $1\times1$ convolution filters to downsample the channel number, which is equivalent to a fully-connected (FC) layer. This directly maps the channel number to the coordinates in an efficient manner. Finally the pose landmarks $\hat{\mathbf{y}}=\{(\hat{a_i},\hat{b_i})| i\in[1,17]\}$ are generated by average pooling from $2\times17\times17$ to $2\times17$ that are the coordinates of 17 human pose landmarks. In a glimpse, the feature extraction of WPNet is achieved by convolutional layers, while the regression function of WPNet is realized by the bottleneck layer with the average pooling.

\begin{table}[]
	\centering
	\caption{The network architecture of WPNet.}\label{tab:network}
	\resizebox{0.4\textwidth}{!}{
		\begin{tabular}{|c|cc|}
			\toprule
			Input   & \multicolumn{2}{c|}{$x_c\in \mathbb{R}^{1\times136\times136}$}                                                                                                                                                                     \\ \midrule
			Block Name    & \multicolumn{1}{c|}{Output Size}            & Parameters   \\ \midrule
			
			Block 1 & \multicolumn{1}{c|}{$64\times136\times136$} & \begin{tabular}[c]{@{}c@{}}$3\times3, 64$\\
			\end{tabular}  
			\\ \midrule
			Block 2 & \multicolumn{1}{c|}{$64\times136\times136$}   & \begin{tabular}[c]{@{}c@{}}$\bigg[\begin{aligned}         & 3\times3, 64 \\         & 3\times3, 64     \end{aligned}     \bigg]\times 3$\end{tabular}   \\ \midrule
			Block 3 & \multicolumn{1}{c|}{$128\times68\times68$}  & \begin{tabular}[c]{@{}c@{}}$\bigg[\begin{aligned}         & 3\times3, 128 \\         & 3\times3, 128     \end{aligned}     \bigg]\times 4$\end{tabular} \\ \midrule
			Block 4 & \multicolumn{1}{c|}{$256\times34\times34$}  & \begin{tabular}[c]{@{}c@{}}$\bigg[\begin{aligned}         & 3\times3, 256 \\         & 3\times3, 256     \end{aligned}     \bigg]\times 6$\end{tabular} \\ \midrule
			Block 5 & \multicolumn{1}{c|}{$512\times17\times17$}  & \begin{tabular}[c]{@{}c@{}}$\bigg[\begin{aligned}         & 3\times3, 512 \\         & 3\times3, 512     \end{aligned}     \bigg]\times 3$\end{tabular} \\ \midrule
			
			Bottleneck & \multicolumn{1}{c|}{$2\times17\times17$}  & \begin{tabular}[c]{@{}c@{}}$1\times1,2$\end{tabular} \\ \midrule
			
			Output  & \multicolumn{2}{c|}{$\hat{y}\in \mathbb{R}^{2\times17}$}                                                                                                                                                                                   \\ \bottomrule
		\end{tabular}
	}
\end{table}

\subsection{Learning Objective}
To train the WPNet, we use the Mean Squared Error (MSE) between the prediction and the annotation generated by the visual pose estimation model as the loss function. The MSE loss is defined by
\begin{equation}
    \mathcal{L}_{MSE}=||\hat{\mathbf{y}}-\mathbf{y}||^2_2.
\end{equation}
Though existing works use pose adjacency matrix~\cite{wang2019can}, we find that using such transformation matrix degrades the performance in the MetaFi. The reason behind it might stem from different granularity of CSI data. As our system can generate 114 subcarriers of CSI data for each pair of antennas, higher resolution data is acquired and thus the task will be more feasible to achieve. It is not necessary to design a more complicated loss function. 
\section{Experiments}
\subsection{Setup}
Our dataset contains data of 13 volunteers (including both men and women). Each volunteer was invited to perform daily-life actions continuously for 6 minutes at different positions, contributing to totally 78 minutes of synchronized video and CSI data. The sampling rates of video and CSI are about 31 Hz and 1000 Hz, respectively, which means one video frame corresponds to 32 CSI packets. We employ 60\% of the data as training set, 20\% as validation set and 20\% as testing set, with the data size 8025, 2676 and 2676, respectively. 

Our WPNet is implemented in PyTorch. We train the model for 50 epochs and employ the stochastic gradient descent (SGD) algorithm to optimize the loss function with the batch size of 32, learning rate of 0.001 and momentum of 0.9. The learning rate follows a multi-step decay rule that was multiplied by 0.5 every 10 epochs.

The widely used Percentage of Correct Keypoint (PCK) in pose estimation~\cite{wang2019can} serves as our evaluation metric, defined by 
\begin{equation}
    PCK_i@a = \frac{1}{N}\sum_{i=1}^NI(\frac{\Vert pd_i - gt_i \Vert_2^2}{\sqrt{rs^2 + lh^2}} \leq a ),
\end{equation}
where $I(\cdot)$ is a logic indicator which outputs 1 while true and 0 while false. $N$ is the amount of test frames, and $i\in\{1,2,\dots,17\}$ is the index of body joint. $a$ is the manually set threshold to adjust the rigidness of the evaluation. $\Vert pd_i - gt_i \Vert_2^2$ represents the Euclidean distance between the predicted results (\textit{pd}) and ground-truth (\textit{gt}), while $\sqrt{rs^2 + lh^2}$ denotes the length between right shoulder ($rs$) and the left hip ($lh$), also termed as the length of torso, which is used to normalize the prediction error.

\begin{table}[]
	\centering
	\caption{Experimental results of MetaFi in the real world. (L. denotes left and R. denotes right)}\label{tab:results}
	\scalebox{0.82}{
		\begin{tabular}{l|cccccc}
			\toprule
			Keypoint   & PCK@5 & PCK@10 & PCK@20 & PCK@30 & PCK@40 & PCK@50 \\ \midrule
			Nose       & 32.21 & 64.72  & 87.71  & 95.03  & 97.87  & 99.40  \\
			L.Eye      & 32.77 & 65.81  & 87.97  & 94.88  & 97.76  & 99.22  \\
			R.Eye      & 33.11 & 66.22  & 88.45  & 95.10  & 98.21  & 99.36  \\
			L.Ear      & 36.62 & 70.63  & 89.28  & 95.44  & 98.28  & 99.36  \\
			R.Ear      & 38.90 & 71.34  & 89.54  & 95.52  & 98.17  & 99.51  \\
			L.Shoulder & 39.42 & 73.73  & 91.22  & 96.60  & 99.07  & 99.70  \\
			R.Shoulder & 36.40 & 72.72  & 91.52  & 96.82  & 99.03  & 99.66  \\
			L.Elbow    & 17.68 & 39.91  & 65.25  & 80.51  & 88.86  & 93.91  \\
			R.Elbow    & 18.24 & 41.85  & 67.04  & 80.12  & 87.18  & 91.89  \\
			L.Wrist    & 12.33 & 31.76  & 52.84  & 63.45  & 68.87  & 74.33  \\
			R.Wrist    & 14.57 & 33.33  & 52.24  & 62.07  & 68.24  & 73.13  \\
			L.Hip      & 43.76 & 82.77  & 98.06  & 99.22  & 99.48  & 99.74  \\
			R.Hip      & 46.26 & 84.27  & 97.72  & 99.14  & 99.63  & 99.78  \\
			L.Knee     & 45.14 & 76.91  & 91.26  & 95.85  & 97.91  & 98.65  \\
			R.Knee     & 43.64 & 77.58  & 91.41  & 96.41  & 97.94  & 98.84  \\
			L.Ankle    & 35.65 & 68.61  & 86.32  & 91.85  & 94.51  & 96.30  \\
			R.Ankle    & 40.99 & 68.95  & 85.99  & 91.97  & 94.54  & 96.15  \\ \midrule
			Average    & 33.39 & 64.18  & 83.16  & 90.00  & 93.27  & 95.23 \\ \bottomrule
		\end{tabular}
	}
\end{table}

\begin{table}[]
	\centering
	\caption{Performance comparison. (WiSPPN* uses different dataset collected by Intel 5300 NIC. WPNet-PAM denotes WPNet trained with PAM loss, the same as WiSPPN.)}\label{tab:comparison}
	\scalebox{0.8}{
		\begin{tabular}{l|cccccc}
			\toprule
			Method   & PCK@5 & PCK@10 & PCK@20 & PCK@30 & PCK@40 & PCK@50 \\ \midrule
			WiSPPN*       & 4.00 & 14.00  & 38.00  & 59.00  & 73.00  & 82.00  \\
			WPNet-PAM    & 19.47 & 48.66  & 77.82  & 87.44 & 91.79  & 94.13 \\
			WPNet & 33.39 & 64.18  & 83.16  & 90.00  & 93.27  & 95.23 \\ \bottomrule
		\end{tabular}
	}
\end{table}

\begin{figure*}[t]
	\centering
	\includegraphics[width=1.0\linewidth]{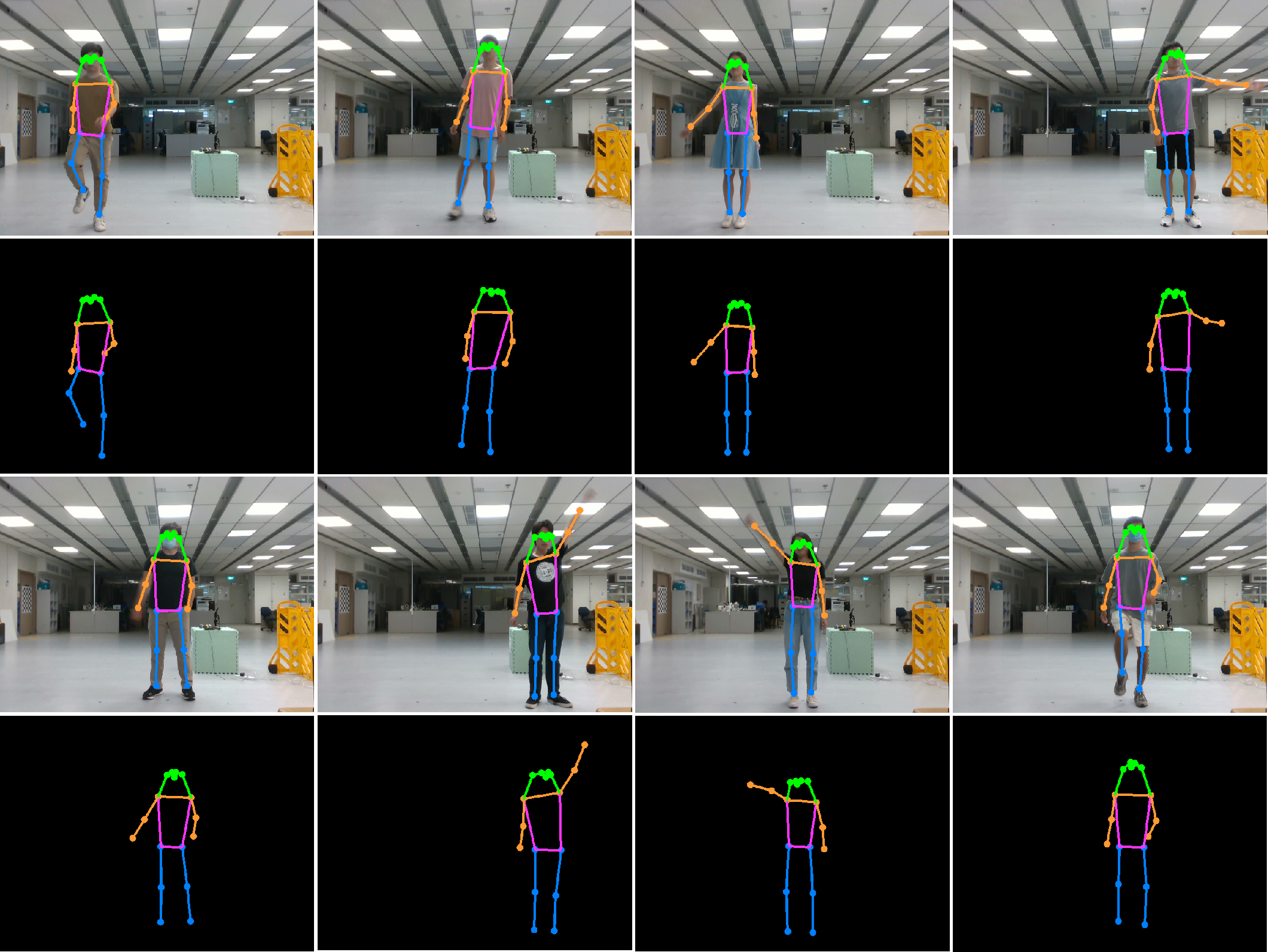}
	\caption{Visualization of the human pose landmarks generated by vision model (up) and WiFi model (down).}
	\label{fig:visualization}
\end{figure*}

\subsection{Results}

We evaluated the performance of WPNet with PCK@5, PCK@10, PCK@20, PCK@30, PCK@40 and PCK@50, with results shown in Table~\ref{tab:results}. From the results, we can observe that our network achieves a good overall pose estimation performance, and the average PCK@50 is 95.23\%. Even the average PCK@5 of our method reaches 33.39\%. The previous work WiSPPN~\cite{wang2019can} only achieves a PCK@50 of 82.00\% based on the Intel 5300 NIC system, whereas our system achieves better PCK@5, PCK@10, PCK@20 and PCK@30 by a large margin. This difference may be caused by the better resolution of CSI in our system. Then we replaced the MSE loss of WPNet to PAM loss that is proposed in WiSPPN, and compared the effect of loss functions. Table~\ref{tab:comparison} shows that our model still outperforms WiSPPN significantly among all 6 evaluation criteria, illustrating that the MSE loss works better in our system.

As for specific body joints, WPNet predicts accurate results for nose, eyes, ears, shoulders and hips with PCK@50 above 99\%. However, the prediction of arms and legs is not satisfactory, especially for the wrists, since the motions of limbs and wrists are subtle. Moreover, the rapid and sophisticated movements of limbs are difficult to distinguish in current CSI sensing. It is assumed that more pairs of antennas will provide better granularity of CSI data and improve the accuracy limbs and wrists, which will be realized and tested in future work.


\subsection{Visualization and Analysis}

Figure~\ref{fig:visualization} demonstrates the comparison of visualization results between HRNet (vision-based solution) and WPNet (ours). We can observe that our WiFi-based method can obtain competitive performance to vision-based methods, inferring positions and various body movements with high accuracy. It is admitted that MetaFi still has certain limitations in the prediction of limbs, including the angle deviation of arms and lack of the bending of legs.

As our MetaFi aims to simulate avatar in smart home, its efficiency also matters. With one NVIDIA RTX 3090 GPU, the inference time for each CSI frame is 0.003s, which means that over 300 frames per second (FPS) can be achieved. Therefore, it is efficient enough for the application of metaverse avatar simulation in smart home.



\subsection{Limitations of MetaFi for Metaverse Applications}
Though the MetaFi attains excellent performance, it still confronts difficulties in real-world metaverse applications. Firstly, the environment dynamics may degrade the performance. Secondly, current WiFi sensing cannot achieve satisfactory accuracy for multi-user scenarios. Thirdly, the method should be deployed conveniently in a different environment. These challenges deserve the future work.

\section{Conclusion}
In this paper, we propose a WiFi-based pose estimation system for metaverse avatar applications in smart home. The system only leverages two COTS WiFi routers for accurate pose estimation, so it is ubiquitous and cost-effective. In MetaFi, the WPNet is designed to extract the spatial-temporal features from CSI data and learn the capacity of pose estimation from advanced computer vision algorithms. The real-world experiments show that our method achieves the PCK@50 of 95.23\%, and the visualization demonstrates the high quality of human pose estimation, which is sufficient for real-time avatar simulation in metaverse.

\bibliographystyle{IEEEtran}
\bibliography{egbib}

\end{document}